# Developing the Path Signature Methodology and its Application to Landmark-based Human Action Recognition

Weixin Yang, Terry Lyons, Hao Ni, Cordelia Schmid, *Fellow, IEEE*, Lianwen Jin, *Member, IEEE*

**Abstract**—Landmark-based human action recognition in videos is a challenging task in computer vision. One key step is to design a generic approach that generates discriminative features for the spatial structure and temporal dynamics. To this end, we regard the evolving landmark data as a high-dimensional path and apply non-linear path signature techniques to provide an expressive, robust, non-linear, and interpretable representation for the sequential events. We do not extract signature features from the raw path, rather we propose path disintegrations and path transformations as preprocessing steps. Path disintegrations turn a high-dimensional path linearly into a collection of lower-dimensional paths; some of these paths are in pose space while others are defined over a multiscale collection of temporal intervals. Path transformations decorate the paths with additional coordinates in standard ways to allow the truncated signatures of transformed paths to expose additional features. For spatial representation, we apply the signature transform to vectorize the paths that arise out of pose disintegration, and for temporal representation, we apply it again to describe this evolving vectorization. Finally, all the features are collected together to constitute the input vector of a linear single-hidden-layer fully-connected network for classification. Experimental results on four datasets demonstrated that the proposed feature set with only a linear shallow network and Dropconnect is effective and achieves comparable state-of-the-art results to the advanced deep networks, and meanwhile, is capable of interpretation.

**Index Terms**—Path signature feature, path disintegration, path transformation, human action recognition.

## 1 Introduction

Human action recognition (HAR) is one of the most challenging tasks in computer vision with a wide range of applications, such as human-computer interaction, video surveillance, behavioral analysis, etc. A vast literature has been devoted to this task in recent years, among which are some informative surveys [1], [2], [3], [4], [5], [6], [7], [8]. An attractive option of HAR is Landmark-based HAR (LHAR) where the object is regarded as a system of correlated labelled landmarks. Johansson's classic moving light-spots experiment [9] demonstrated that people can detect motion patterns and recognize actions from several bright spots distributed on the body, which has stimulated research on pose estimation and LHAR [10], [11], [12]. Different from skeleton-based HAR (SHAR), LHAR, using no knowledge of skeletal structure, is flexible to extend to any landmark data streams with no explicit physical structures, e.g. traffic or people flow.

Although many solutions have been proposed to address the challenge of LHAR, the problem remains unsolved due to two main challenges. First, there is the problem of designing reliable discriminative features for spatial structural representation, and second of modelling the temporal dynamics of motion. In this paper, the path signature feature (PSF) is used and refined as an expressive, robust, non-linear, and interpretable feature set for spatial and temporal representation of LHAR.

The path signature, which was initially introduced in rough paths theory as a branch of stochastic analysis, has been successfully applied to many machine learning tasks. Most existing work can be devided into two categories: sliding-window-based and global-based. In the sliding temporal window approach [24], [25], [26], [27], [28], [29], [63], signatures of small paths are extracted and embedded into multi-channel feature maps as input of a CNN. The signatures herein are merely local descriptors from which the deep models are then trained to learn hierarchical representation. The global-based approaches combine all the cues into a high-dimensional path to compute high-level signatures over the whole time interval [30], [62] or low-level signatures over subsampling intervals [61]. They are straightforward but not efficient for high dimensional or spatio-temporal data.

To represent spatial pose, most methods [12], [19], [23], [41], [46], [47], [51], [92] used predefined skeletal structures. The connections distributed on a physical body are intuitive spatial constraints but not necessarily the crucial ones to distinguish actions. The connections discarded by imposing a skeletal structure could contain valuable non-local information. To solve this, hand-designed features [31], [42], [43], [44] were employed, but they are limited to encode non-linear dependencies. In this paper, we pro-

- *Weixin Yang is with Mathematical Institute, University of Oxford, UK and was with College of Electronic and Information Engineering, South China University of Technology, China. E-mail: wxy1290@163.com.*
- *Terry Lyons is with Mathematical Institute, University of Oxford, UK, and Alan Turing Institute, UK. E-mail: tlyons@maths.ox.ac.uk.*
- *Hao Ni is with Department of Mathematics, University College London, UK, and Alan Turing Institute, UK. E-mail: ucahhni@ucl.ac.uk.*
- *Cordelia Schmid is with Thoth project-team, Inria Grenoble Rhone-Alpes, Laboratoire Jean Kuntzmann, France. E-mail: cordelia.schmid@inria.fr.*
- *Lianwen Jin is with College of Electronic and Information Engineering, South China University of Technology, China. E-mail: lianwen.jin@scut.edu.cn.*

pose to localize a pose by disintegration into a collection of $m$-node sub-paths. The signatures of these paths encode non-local and non-linear geometrical dependencies.

To model temporal dynamics, hand-designed local descriptors [31], [44] were popular, but it is difficult to encode complex spatio-temporal dependences in these. Recently, recurrent neural networks (RNN) [16], especially long short-term memory (LSTM) [17], have gained increasing popularity in handling sequential data, including human actions [18], [19], [20], [21]. In particular, a variation of LSTM [22], [23] succeeded in simultaneously exploring both spatial and temporal information. These deep models play a vital role in feature representation and achieve state-of-the-art performance, but the features learned by them are not as interpretable as hand-designed features. In this paper our temporal disintegration turns the original paths into hierarchical paths, from which the signatures encode multi-scale dynamical dependencies. Moreover, our path transformations decorate the paths with additional coordinates to allow signatures to expose additional valuable features.

To build the spatial and temporal representation, in each frame the spatial PSFs are extracted from the localized paths obtained by pose disintegration. In the clip, the evolution of each spatial feature along the time axis constitutes a spatio-temporal path. After path transformations and temporal disintegration, the temporal PSFs are then extracted from the spatio-temporal paths. Finally, the collection of all the features forms the input vector of a linear single-hidden-layer fully-connected network for classification. To extensively evaluate the effectiveness and flexibility of our method, several datasets (i.e., JHMDB [31], SBU [32], Berkeley MHAD [33], and NTURGB+D [21]) collected by different acquisition devices were used for experiments. Using our feature set and only a linear shallow net, we achieve comparable results to the advanced deep learning methods. Moreover, we took a further step toward understanding human actions by analyzing the PSFs and the linear classifier.

Our major contributions lie in four aspects:

1. PSFs are adopted and refined for LHAR with interpretations, proofs, experiments, and discussions of their properties and advantages.
2. Pose disintegration is proposed for non-local spatial dependencies, and temporal disintegration is proposed for multiscale temporal dependencies.
3. Path transformations, decorating the original paths with additional coordinates, are proposed to allow signatures to expose additional features.
4. Using signature-based spatio-temporal representation and only a linear shallow net, we achieve comparable state-of-the-art results to those with deep models. Meanwhile, this interpretable pipeline facilitates the understanding of HAR.

## 2 Related Work

### 2.1 Landmark-based human action recognition

A human body can be regarded as an articulated system composed of joints that evolve in time [35]. For recent surveys of LHAR, we refer the reader to [8], [36], [37].

Approaches for LHAR can be categorized into two classes: joint-based and part-based. The joint-based ones regard the human body as a set of points and attempt to capture the correlation among body joints by using the motion of 3D points [38], [39], measuring the pairwise distances [31], [40], [41], [42], [43], [44], or using the joint orientations [45]. On the other hand, the part-based approaches focus on connected segments of the human skeleton. They group the body into several parts and encode these parts separately [46], [47], [48], [49], [50], [51], [52]. Some methods in this category represent a pose by means of the geometric relations among body parts, for examples, [46], [47] employed quadruples of joints to form a new coordinate system for representation, and [12] considered measurements of the geometric transformation from one body part to another. Some methods assume that certain actions are usually associated with a subset of body parts, so they aim to identify and use the subsets of the most discriminative parts of the joints.

Given the recent success of deep learning frameworks, some works aim to capture correlation among joint positions using CNNs [53], [54], [55], [56]. In [53], the input feature maps of a CNN were joints colored according to their sequential orders, body parts, or velocity, while in [54] and [55], the CNN's inputs were the collection of hand-designed local features. Since human actions are usually recorded as video sequences, it is natural to apply RNNs or LSTMs. HBRNN [19] and Part-aware LSTM [21] contained multiple networks for different groups of joints. Zhu et al. [18] proposed a deep LSTM to learn the co-occurrence of discriminative joints using a mixed-norm regularization term in the cost function. By additional new gating to the LSTM, the Differential LSTM [20] is able to discover the salient motion patterns, and [22], [23] achieved robustness to noise. It is noteworthy that the spatio-temporal RNNs in [22], [23] concurrently encoded both spatial and temporal context of actions within a LSTM. Liu et al. [86] used an attention-based LSTM to iteratively select informative keypoints for recognition. Zhang et al. [87] used a multilayer LSTM to fuse several simple geometric features for recognition. The CNN and RNN frameworks achieve high accuracy on most SHAR datasets, showing excellent feature learning capability.

### 2.2 Path signature feature (PSF)

Rough path theory is concerned with capturing and making precise the interactions between highly oscillatory and non-linear systems [57]. The essential object in rough path theory, called the path signature, was first studied by Chen [58] whose work concentrates on piecewise regular paths. More recently, the path signature has been used by Lyons [34] to make sense of the solution to differential equations driven by very rough signals. It was extended by Lyons' theory from paths of bounded variation [34] to rough paths of finite $p$-variation for any $p \geqslant 1$ [59].

Some successful applications of the PSF have been made in the fields of machine learning, pattern recognition and data analysis. For financial data, useful predictions can be made with only a small number of truncated

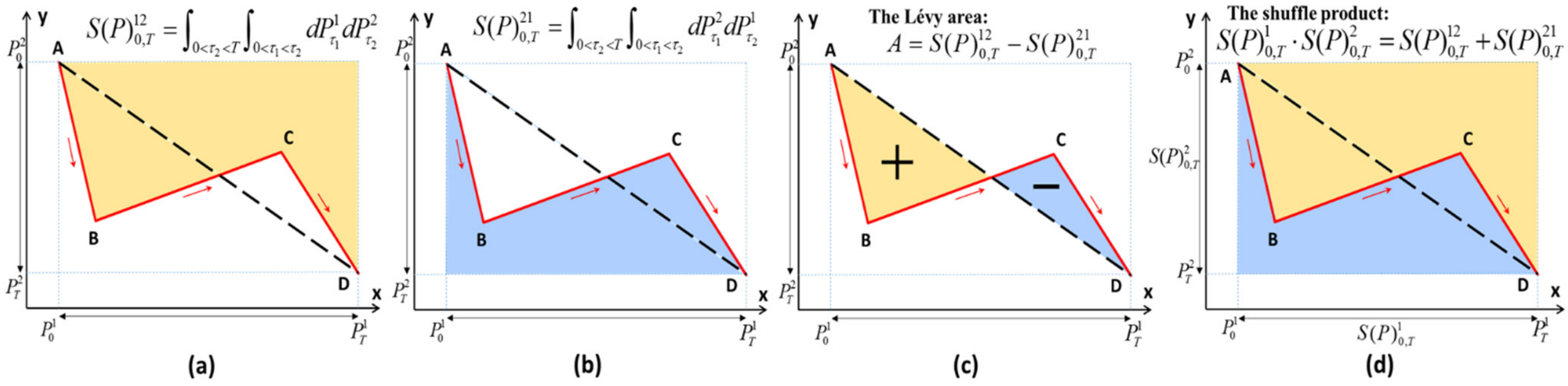

Fig. 1. The geometric intuition of the PSF of a 2D path. The path in red moves from A to D over the time interval [0, T]. The dashed line is the chord connecting the endpoints. Panels (a) and (b) depict two terms of the 2-fold iterated integrals of the path, (c) is the Lévy area enclosed by the path and its chord, and (d) is a demonstration of the shuffle product identity.

PSFs [30], [91]. In [60], a signature-based kernel framework for hand movement classification was presented. Moreover, PSFs were used on self-reported mood data to distinguish psychiatric disorders [61]. One of the most notable applications of using PSFs is handwriting understanding. Diehl [62] used iterated integrals of a handwritten curve for recognition and found that some linear functions of the PSF satisfy rotation invariance. Graham [63] used the sliding-window-based PSF as feature maps of a CNN for large-scale online handwritten character recognition, based on which he won the ICDAR2013 competition [64]. Inspired by this, Xie et al. [26], [27] extended the method to handwritten text recognition. Yang et al. [28], [29] explored the higher-level terms of the PSF for text-independent writer identification which requies subtle geometric features. Overall, these applications demonstrate the value of the PSF as an effective and informative vector representation for sequential data.

## 3 Path Signature

### 3.1 Definition and geometric interpretation

The rigorous introduction of the path signature as a faithful description or feature set for un-parameterized paths can be found in [57], [65], [66], [67], so in this paper we present it in a practical manner.

A $d$-dimensional path or stream of timestamped events $P$ over the time interval $[0,T] \subset \mathbb{R}$ can be interpolated to a continuous map $P:[0,T] \to \mathbb{R}^d$. The coordinates of $P$ at time $\tau$ are $P_\tau = (P^1_\tau, P^2_\tau, \dots, P^d_\tau)$. Consider the simplest case when $d = 1$. The path $(P^1_\tau)$ is a real-valued path for which the path integral is defined as

$$S(P)^1_{0,T} = \int_{0<\tau\le T} dP^1_\tau = P^1_T - P^1_0, \tag{1}$$

which is the increment of this 1-dimensional path over the whole time interval and is called the 1-fold iterated integral. We emphasize that $S(P)^1_{0,\tau}, 0<\tau\le T$ is also a real-valued path w.r.t $\tau$. The 2-fold iterated integral is

$$S(P)^{11}_{0,T} = \int_{0<\tau_2\le T} S(P)^1_{0,\tau_2}\, dP^1_{\tau_2} = \tfrac{1}{2}(P^1_T - P^1_0)^2, \tag{2}$$

which is proportional to the square of the increment. Again, $S(P)^{11}_{0,\tau}$ is a real-valued path, so if we continue recursively, the $k$-fold iterated integral of $P$ is

$$\begin{aligned} S(P)^{11\dots1}_{0,T} &= \int_{0<\tau_k\le T}\cdots\int_{0<\tau_2\le\tau_3}\int_{0<\tau_1\le\tau_2} dP^1_{\tau_1} dP^1_{\tau_2}\dots dP^1_{\tau_k} \\ &= \tfrac{1}{k!}(P^1_T - P^1_0)^k, \end{aligned} \tag{3}$$

which is proportional to the increment to the power of $k$.

Now, when $d=2$, the 1-fold iterated integral of the path $\{P^1_\tau, P^2_\tau\}$ has 2 elements

$$S(P)^1_{0,T} = \int_{0<\tau\le T} dP^1_\tau = P^1_T - P^1_0, \tag{4}$$

$$S(P)^2_{0,T} = \int_{0<\tau\le T} dP^2_\tau = P^2_T - P^2_0. \tag{5}$$

Each element is the increment of the path on the corresponding axis over the time interval $[0,T]$. They denote the displacement of the given path. The 2-fold iterated integral of this 2D path contains $d^2 = 2^2$ elements

$$S(P)^{11}_{0,T} = \int_{0<\tau_2\le T}\int_{0<\tau_1\le\tau_2} dP^1_{\tau_1} dP^1_{\tau_2} = \tfrac{1}{2!}(P^1_T - P^1_0)^2, \tag{6}$$

$$S(P)^{22}_{0,T} = \int_{0<\tau_2\le T}\int_{0<\tau_1\le\tau_2} dP^2_{\tau_1} dP^2_{\tau_2} = \tfrac{1}{2!}(P^2_T - P^2_0)^2, \tag{7}$$

$$S(P)^{12}_{0,T} = \int_{0<\tau_2\le T}\int_{0<\tau_1\le\tau_2} dP^1_{\tau_1} dP^2_{\tau_2}, \tag{8}$$

$$S(P)^{21}_{0,T} = \int_{0<\tau_2\le T}\int_{0<\tau_1\le\tau_2} dP^2_{\tau_1} dP^1_{\tau_2}. \tag{9}$$

We note that the first two elements are the same as (2) in the 1-dimensional case. For the other two elements, the geometric intuitions are the areas shown in Fig. 1(a) and Fig. 1(b). Together they represent the Lévy area [65] depicted in Fig. 1(c). The Lévy area, which is a signed area enclosed by the path and the chord connecting the endpoints, can be expressed by

$$A_{0,T} = S(P)^{12}_{0,T} - S(P)^{21}_{0,T}. \tag{10}$$

The sign of the area depends on the sign of the winding number of the path moving around it [68]. The interpretation of the $k$-fold iterated integral ($k > 2$) of a 2D path is not trivial, so it is not included here. By analogy, for a 3D path, the 1-fold, 2-fold, and 3-fold iterated integrals are units of displacement, area, and volume respectively.

In general, for a path in $\mathbb{R}^d$, the superscript of the $k$-fold iterated integral, which describes the order of integration, is a multi-index $(i_1, i_2, \dots, i_k) \in \{1,\dots,d\}^k$. Therefore, the $d^k$ elements of the $k$-fold iterated integral of a $d$-dimensional path can be generally expressed as

$$S(P)^{i_1,i_2,\dots,i_k}_{0,T} = \int_{0<\tau_k\le T}\cdots\int_{0<\tau_2\le\tau_3}\int_{0<\tau_1\le\tau_2} dP^{i_1}_{\tau_1} dP^{i_2}_{\tau_2}\dots dP^{i_k}_{\tau_k}. \tag{11}$$

Then the signature of a path $P$ over the time interval $[0,T]$ is the collection of all the iterated integrals of $P$:

$$\begin{aligned} S(P)_{0,T} = (\, & 1, S(P)^1_{0,T}, \dots, S(P)^d_{0,T}, \\ & S(P)^{1,1}_{0,T}, \dots, S(P)^{1,d}_{0,T}, S(P)^{2,1}_{0,T}, \dots, S(P)^{d,1}_{0,T}, \dots, S(P)^{d,d}_{0,T}, \\ & \dots, S(P)^{1,1,\dots,1}_{0,T}, \dots, S(P)^{i_1,i_2,\dots,i_k}_{0,T}, \dots, S(P)^{d,d,\dots,d}_{0,T}, \dots\,), \end{aligned} \tag{12}$$

where the zeroth term is conventionally set to 1. Since the signature is defined on top of all the possible indices of finite length, the number of elements in the signature is infinite. In practical use we usually consider the signature truncated at a certain level $n$ written as

$$S_n(P)_{0,T} = (1, S(P)^{1}_{0,T}, \ldots, S(P)^{i_1,i_2,\ldots,i_n}_{0,T}, \ldots, S(P)^{d,d,\ldots,d}_{0,T}), \quad (13)$$

of which the dimensionality is $\varphi(d,n) = (d^{n+1}-1)(d-1)^{-1}$. The elements of the truncated signature are taken as features (i.e., PSF) encoding the informative geometric properties of sequential data in applications in machine learning. For the feature set, the 0-th term (i.e., a constant value set to 1) is optional, so the dimension can be reduced to

$$\varphi'(d,n) = (d^{n+1}-d)(d-1)^{-1}. \quad (14)$$

For the 1-dimensional case ($d$ = 1), the feature dimension is exactly equal to $n$ (excluded the zeroth term) according to (1), (2), and (3).

### 3.2 Calculation of the signature for a discrete path

Although the path signature is initially defined for continuous paths with bounded variation, it is easily extended to discrete paths by linear interpolation [95]. The signature is canonical and does not depend on the choice of timescale used for the interpolation.

Computing the signature of a piecewise linear path does not require integrals. For each line segment of the path, the elements of its signature are given by

$$S(P)^{i_1,i_2,\ldots,i_k}_{\tau,\tau+1} = \frac{1}{k!}\prod_{j=1}^{k}(P^{i_j}_{\tau+1} - P^{i_j}_{\tau}), \quad (15)$$

where $P^{i_j}_{\tau}$ is the $i_j$-th coordinate value of path $P$ at time $\tau$. For the entire path, Chen's identity [58] states that for any time stamps ($s$, $t$, $u$) satisfying that $s < t < u$, we have

$$S(P)^{i_1,i_2,\ldots,i_k,\ldots,i_n}_{s,u} = \sum_{k=0}^{n} S(P)^{i_1,i_2,\ldots,i_k}_{s,t} S(P)^{i_{k+1},i_{k+2},\ldots,i_n}_{t,u}. \quad (16)$$

This implies that the signature of the entire path can be calculated from the signatures of its pieces.

We recommend the two open-source python software libraries, *esig* (on PyPi), derived from the *CoRoPa* C++ library [69], and *iisignature* [70]. Both allow fast computation of the path signature. With further optimized code and a good multicore CPU and MKL BLAS an avarage computation time of 0.625 seconds can be achieved for the truncated signature up to level 4 of a 60D path (the dimension of its signature is more than 13 million). Based on CUBLAS on a Nvidia GP100 with no custom cuda kernel, this can be decreased to 0.055 seconds.

### 3.3 Properties of the path signature

#### *3.3.1 Uniqueness*

It is proved that the path signature determines a path up to tree-like equivalence (this notion is introduced in [34]). A tree-like section in a path is a section where the trajectory exactly retraces itself. Tree-like sections are common in real-world data streams, such as some periodic human actions like clapping or jumping in place. Note that a path with a monotone dimension, such as time, has no tree-like sections.

#### *3.3.2 Invariance under translation*

The signature computed by (11) or (15) is invariant under translation of the paths, which has the practical advantage of avoiding complex recentering normalization.

#### *3.3.3 Invariance under reparameterization*

A reparameterization of a path is a continuous, non-decreasing substitution for the parameterized variable of a path. It changes the speed of recording of the path. Human actions are largely invariant under changing the speed of the action or the frame rate of the video. The ease with which the signature can completely filter out these changes in the representation is a major advantage for machine learning, substantially reducing the dimensionality of the feature set needed for action classification. The use of the path signature, with its fixed-dimensional feature set, can help the classifer to recognize the same action performed or sampled at different speeds. We refer the reader to [57], [67] for a detailed proof of the invariance of the path signature under reparameterization.

#### *3.3.4 The generic nonlinearity of the signature*

The shuffle product identity [65] states that the product of two lower-level signature coefficients can be expressed as a linear combination of some higher-level coefficients. For instance, for the two-dimensional case in section 3.1, we can easily derive the following equation from Fig. 1(d),

$$S(P)^{1}_{0,T} \cdot S(P)^{2}_{0,T} = S(P)^{12}_{0,T} + S(P)^{21}_{0,T}. \quad (17)$$

In other words, the nonlinear behavior in terms of lower level terms can be expressed by linear combination of higher-level terms. Therefore, when we incorporate the higher-level terms into the feature representation, we automatically include more nonlinear prior knowledge in our feature set. If the introduced nonlinearity is sufficient, we need only linear classifiers to distinguish the targets.

#### *3.3.5 Fixed dimension under length variations*

Another practical property of the path signature is that the dimension of the PSF extracted from the entire path depends on the truncation level of the signature and the intrinsic dimension of the path but is independent of the (sampled) length of the path, as described in (14). For human action recognition, the durations of actions are variable. The use of the path signature allows us to extract a feature vector with fixed dimension and use it with classification methods which require a fixed-length input.

## 4 PATH DISINTEGRATIONS AND TRANSFORMATIONS

The principled and robust representation of unparameterized paths, along with the convenience of reducing polynomial functions on the space of paths to linear ones (which establishes their universality) provide the core motivations for using signatures as features. One can always take the signature of a raw path to remove any dependence on parameterization or translation, but sometimes it is prudent to apply path disintegrations or path transformations as preprocessing to improve the efficiency and effectiveness of PSFs. The disintegrations turn a path into a collection of subpaths while the transformations turn a path into a higher-dimensional path.

## 4.1 Path disintegrations

### 4.1.1 Pose Disintegration

In many cases, non-local clues are informative and straightforward, for instance, the non-local displacement between two hand points is a key feature for the action of clapping. To exploit both local and non-local clues in pose, we propose pose disintegration. Landmarks that are labelled with corresponding body parts have no inherent order, so a predefined priority order is randomly chosen and fixed -- different random choices of initial order yield comparable results in preliminary experiments. The pose is then regarded as an ordered collection of points in $\mathbb{R}^d$. Our pose disintegration localizes the pose into all possible subposes containing $m$ points. Connecting the $m$ points in each subpose in the inherited order forms a unique $m$-node sub-path that visits each point once. We end up with a collection of sub-paths which do not need to be parts of physical body and are available for further path transformations or signature extractions.

We consider that functions on a pose can be approximated by functions on the piecewise linear localized paths of its subposes. For convenience, one can view these functions as linear functions in the signatures of its localized paths. The terms of the first two levels of signatures cover the displacement and the area information similar to the traditional hand-designed features [31], [44], while the higher-level terms capture more non-linear features. For a pose with $N$ joints, the dimension of the signatures of its localized paths is $C_N^m \cdot \varphi'(d,n)$, where $m$ is the number of points in a subpose, $d$ is the dimension of the path, and $n$ is the truncated signature level. The selections of these parameter values are highly correlated and associated with the uniqueness of the paths. According to [93], any piecewise linear paths in $\mathbb{R}^d$, consisting of at most $m = d+1$ points, can be uniquely recovered from the signature at the third level. A larger $m$ brings semantically high-level components but requires a larger $n$ for the path uniqueness [94], which exponentially increases the feature dimension according to (14), and means less shareability and more sub-paths. The number of $m$-node sub-paths is in line with Pascal's triangle and increases along with $m$ ($m \le N/2$). To avoid feature set of very large dimension, $m \le 3, n = 3$ for $d = 2$ and $m \le 4, n = 3$ for $d = 3$ are suggested. Beyond the signature level $n$ required for the unique recovery of a path, the non-linearity (as described in section 3.3.4) of the extra high-level terms may still contribute to facilitate the training of the model until the dimensionality of the feature set becomes impractical.

### 4.1.2 Temporal Disintegration

Temporal disintegration is based on the basic theory of the path signature which suggests that for long-range paths, low-level terms of signatures on all intermediate length time intervals can be far more efficient than signatures of high levels over the whole time interval [65]. Therefore, instead of extracting the PSF over the whole time interval, the dyadic path signature features (DPSF) [71] split the entire interval into small intervals with a dyadic hierarchical structure and then extracts PSF over each small interval. Given a path over the whole time interval $[0,T] \subset \mathbb{R}$, the $j$-th dyadic level of the hierarchical structure is the collection of subintervals $[iT/2^j, (i+1)T/2^j], i \in [0, 2^j - 1], j \in \mathbb{N}$. Note that the 0-th dyadic level contains exactly the whole path. The DPSF over long, medium, and short time intervals describes multi-scale dynamical activities more efficiently than the PSF over the entire interval, which requires higher-level terms to capture local dependencies.

Slightly different from the hierarchical structure in [71] which may break the events that occur near the conjunctive time stamps $\{iT/2^j \mid i \in [1, 2^j - 1], j \in \mathbb{N}^+\}$, we consider an overlapping version over the time intervals $[iT/2^{j+1}, (i+2)T/2^{j+1}]$, $i \in [0, 2 \cdot (2^j - 1)]$. The overlapping DPSF is expected to supplement the original DPSF with additional local details. Its dimension is

$$\hat{\varphi}(h,d,n) = (2^{h+1} - h - 2) \cdot \varphi'(d,n), \tag{19}$$

where $h \in \mathbb{N}^+$ is the number of the hierarchical level. The selection of $h$ is a tradeoff between improving efficiency and introducing local noise over finer intervals.

## 4.2 Path transformations

### 4.2.1 Time-incorporated transformation

The signature is invariant under parameterization, but in many situations, one would like to keep the dependence on time. Adding a monotone increasing time dimension is adopted to encode motion speed. The signature of a time-incorporated path contains two parts: time-independent (TI) and time-dependent (TD). The TI part is exactly the signature of the original path, so its integration order is

$$i_1, i_2, \ldots, i_k \in \{1, \ldots, d\}. \tag{20}$$

The TD part is related with time. Its integration order is

$$i_1, i_2, \ldots, i_k \in \{1, \ldots, d+1\}, \exists m \in [1,k], i_m = d+1, \tag{21}$$

which means each term of the signature in TD is an integral along the time dimension at least once. Given the truncated signature level $n$, the dimensionality of the TD part is $\varphi'(d+1,n) - \varphi'(d,n)$. The signature of the original path filters out the information about the speed of motion and the sampling rate but the signature of the time-incorporated path allows us to select one and suppress the other according to significance to the classification.

### 4.2.2 Invisibility-reset transformation

The signature capturing relative position information is invariant under translation. Given that the absolute position may be essential for some scenarios (e.g., HAR under static CCTVs), we propose the invisibility-reset transformation of a path to retain the absolute position information in signatures. For a path $P$ in $\mathbb{R}^d$ within the interval $[0,T]$, we add two time steps $T$+1 and $T$+2 with value $P_T$ and $\mathbf{0}$ respectively at the end of $P$ and add a visibility dimension $v$ with values 1 in $[0,T]$ and 0 in $(T, T+2]$. In other words, the invisibility-reset path $P_{IR}$ in $\mathbb{R}^{d+1}$ first becomes invisible at time $T$+1 and then is reset to the origin at $T$+2. According to (15) and (16), we have

$$S(P_{IR})_{0,T+2}^{i_1,i_2,\ldots,i_k,v} = -S(P)_{0,T}^{i_1,i_2,\ldots,i_k}, \; i_1, i_2, \ldots, i_k \in \{1, \ldots, d\} \tag{22}$$

which means certain terms in $S(P_{IR})$ encode the relative positions as in $S(P)$. Moreover, the terms of the first lev-

TABLE 1
PROPOSED FEATURES FOR LANDMARK-BASED HUMAN ACTION RECOGNITION

| # of joints | Spatial structural features (in each frame) | Temporal dynamical features (along the time axis) |
|---|---|---|
| 1 (a single joint) | **S-J**: The $d$-dimensional coordinates of each of the predefined $N$ joints are incorporated. | **T-J-PSF**: The temporal PSF over the evolution of each joint up to signature level $n_{TJ}$ is extracted. |
| 2 (joint pair) | **S-P-PSF**: The PSF over each pair of joint up to signature level $n_{SP}$ is extracted. | **T-S-PSF**: The evolution of each dimension of spatial PSF is treated as path, over which the temporal PSF up to signature level $n_{TS}$ is extracted. |
| 3 (joint triple) | **S-T-PSF**: The PSF over each triple of joint up to signature level $n_{ST}$ is extracted. | |

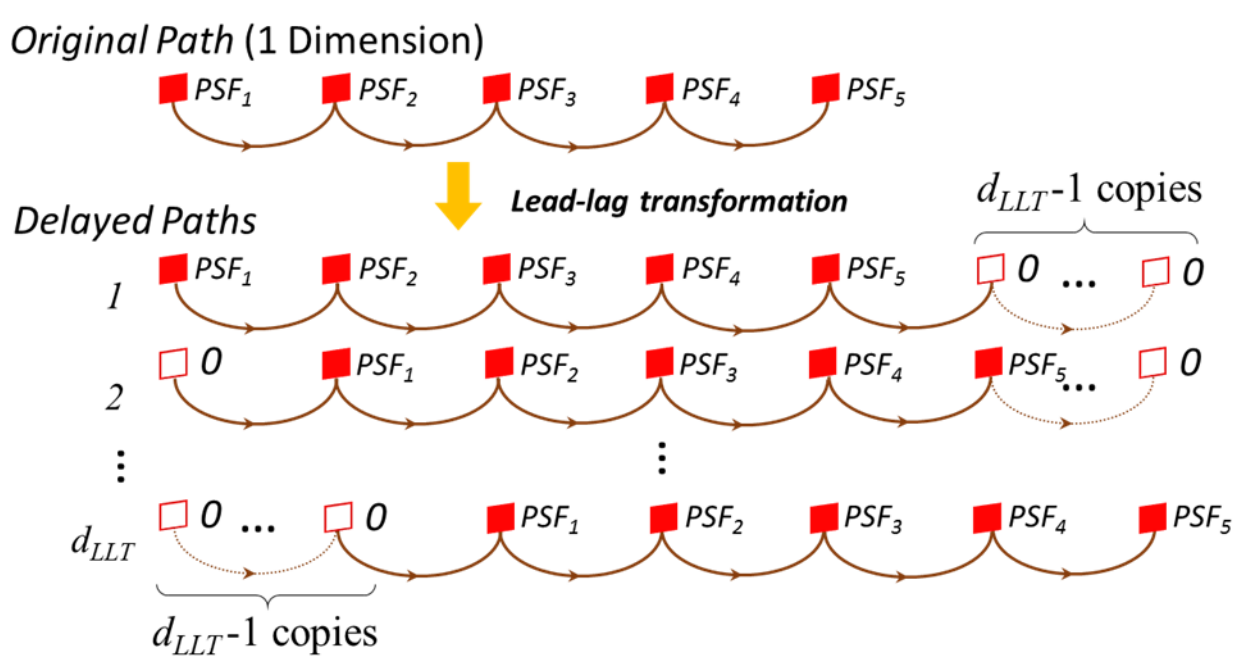

Fig. 2. The illustration of multi-delayed lead-lag transformation. The dimension of lead-lag paths is $d_{LLT}$. The delayed paths are padded with zeros to ensure a fix length for each dimension.

el of $S(P_{IR})$, given by

$$S(P_{IR})^{i_1}_{0,T+2} = -P_0^{i_1}, \; i_1 \in \{1,\dots,d\}, \tag{23}$$

are the absolute position of the initial point. This simple transformation retains different position information in signatures and thus allows the network to select one and suppress the other according to significance to the task.

### 4.2.3 Multi-delayed lead-lag transformation

The lead-lag transformation proposed in [30], [66], [95] is designed to explore the quadratic cross-variation between the original path and its delayed path. To extend its capability to describe long-term dependencies of sequential events, our modified lead-lag transformation, as shown in Fig. 2, is constructed by the original path and its multiple delayed paths (instead of one delayed path in [30]). We denote the dimension of a lead-lag transformed path as $d_{LLT}$. The signatures of lead-lag paths with smaller $d_{LLT}$ encode short-term dependencies, while those with larger $d_{LLT}$ explore more long-term dependencies.

## 5 FEATURE EXTRACTION FOR HUMAN ACTION RECOGNITION

Our proposed feature set for LHAR, which we describe in this section, is outlined in Table 1.

### 5.1 Spatial structural features

First of all, the basic information describing the spatial structure is the $d$-dimensional coordinates of each of the $N$ joints of the body. The keyword **S-J** denotes the spatial coordinate values of the joints. The dimension of this part is $D_{SJ} = N \cdot d$ for each sampled frame.

To encode the spatial context, we use pose disintegration with $m = 2$ and $m = 3$, which means joint pairs and joint triples are used as illustrated in Fig. 3. The signatures of each of these subpaths are computed to model the spatial constraints in each frame. The spatial PSF of joint pairs and joint triples are denoted by **S-P-PSF** and **S-T-PSF** respectively. If the truncation level of the signature of pairs and triples are $n_{SP}$ and $n_{ST}$ respectively, then the feature dimensions per frame are $D_{SP} = C_N^2 \cdot \varphi'(d, n_{SP})$ and $D_{ST} = C_N^3 \cdot \varphi'(d, n_{ST})$ respectively. Finally, the spatial features from all sampled frames are extracted and collected. The dimension of **S-P-PSF** and **S-T-PSF** per frame is denoted by $D_S = D_{SP} + D_{ST}$.

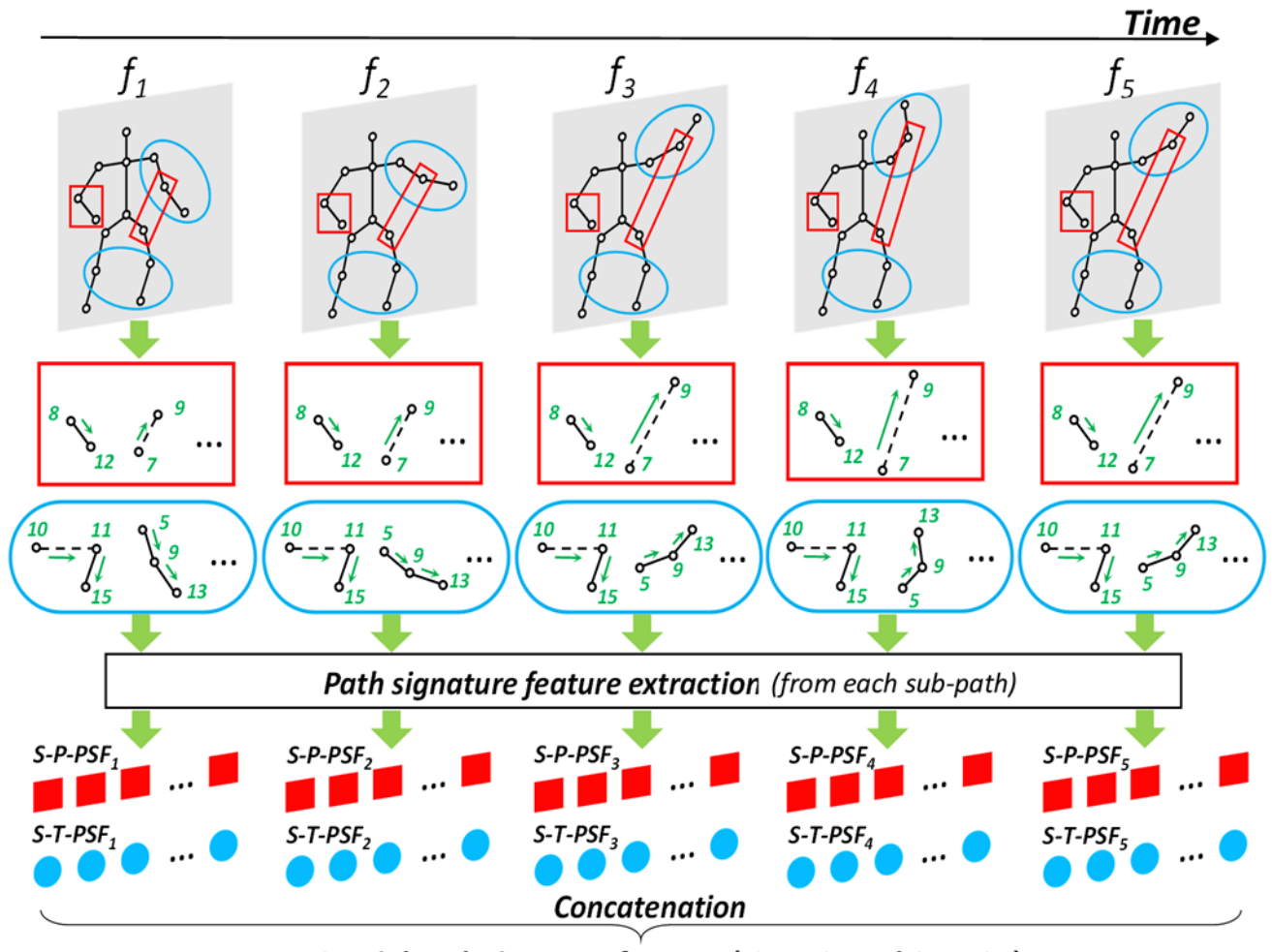

Fig. 3. The illustration of spatial feature (**S-P-PSF** and **S-T-PSF**) extraction. Note that we predefine the priority order of all the N joints (N = 15 in this figure). The red quadrangles denote the feature extraction of joint pairs, while the blue ellipses denote that of joint triples. All possible pairs and triples of joints are considered.

### 5.2 Temporal dynamical features

Inspired by the works in [22], [23] which jointly learned the spatial and temporal contexts in a variant of LSTM, we suggest that the dynamics of landmark-based human action can be described by the evolution of spatial context. The spatial context herein are the features we extracted in section 5.1, although other spatial features can be used alternatively. From these, we are going to extract two kinds of temporal features **T-J-PSF** and **T-S-PSF**.

The **T-J-PSF**, illustrated in Fig. 4, is the temporal PSF of the evolution of each joint along the time. The evolution of each joint is naturally a time-sequence, so we consider its time-incorporated transformation. For $N$-joint bodies in $\mathbb{R}^d$, the dimension of **T-J-PSF** is $D_{TJ} = N \cdot \varphi'(d+1, n_{TJ})$, where $n_{TJ}$ is the truncation level of the signature.

Since each dimension of the spatial contextual features (**S-P-PSF** and **S-T-PSF**) characterizes one particular spatial constraint of a pose, the evolution of this spatial constraint along the time forms a spatio-temporal path which delivers temporal constraints of a motion. The temporal PSF of these spatio-temporal paths is denoted by **T-S-PSF** and illustrated in Fig. 5. Since the signature of a spatio-temporal path (i.e., a 1D path) is just the increments to a certain power, the multi-delayed lead-lag transformation is applied to each path to enrich the PSF with cross-variations among events of the path. If the truncation level of the signature is $n_{TS}$ and the dimension of the lead-lag paths is $d_{LLT}$, the dimension of **T-S-PSF** from all spatio-temporal paths is $D_{TS} = D_S \cdot \varphi'(d_{LLT}, n_{TS})$. Considering there might exist complicated or long-range actions, the temporal disintegration in section 4.1.2 can be applied. If so, the dimensions are $D_{TJ} = N \cdot \hat{\varphi}(h_{TJ}, d+1, n_{TJ})$ and $D_{TS} = D_S \cdot \hat{\varphi}(h_{TS}, d_{LLT}, n_{TS})$ where $h_{TJ}$ and $h_{TS}$ are the corresponding hierarchical levels.

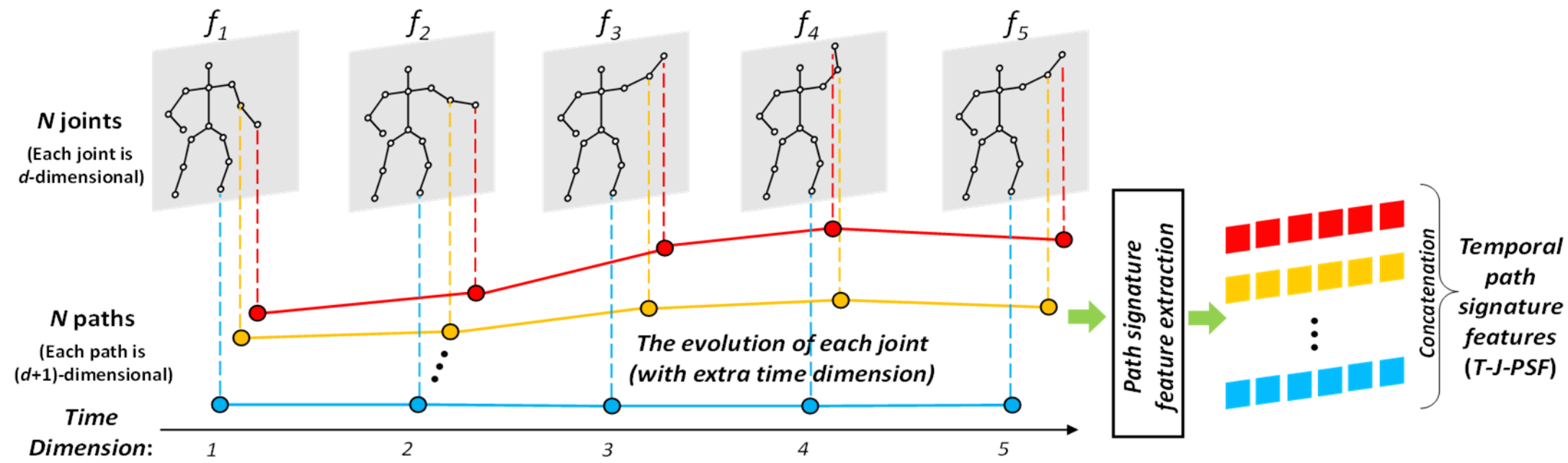

Fig. 4. Illustration of temporal features extracted from the evolution of each corresponding joint (**T-J-PSF**).

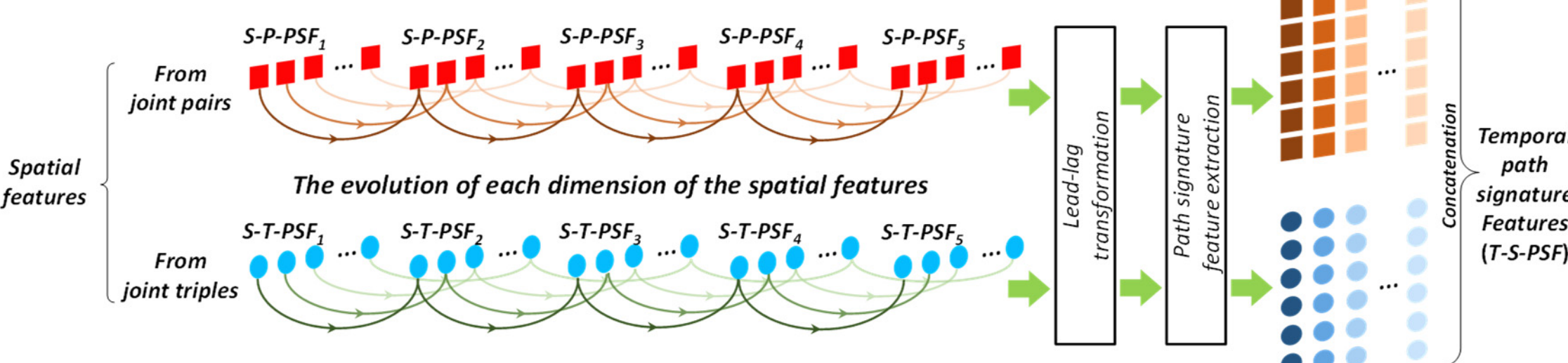

Fig. 5. Illustration of temporal features extracted from the evolution of spatial context (**T-S-PSF**). Each dimension of the spatial features is treated equally and individually.

The dimension of all temporal PSFs is $D_T = D_{TJ} + D_{TS}$. Finally, the total dimension of spatial and temporal features per clip is $D = M \cdot (D_{SJ} + D_S) + D_T$, where $M$ is the number of sampled frames. Moreover, the spatial components can be covered by the temporal PSFs extracted from invisibility-reset paths which require no sampling step.

## 6 Experimental Results and Analysis

### 6.1 Datasets

Monocular videos recorded by 2D cameras are capable of collecting spontaneous actions, but their sensitivity to viewpoint variations and occlusions makes recognition a difficult task [1]. Intuitively, human body is general in 3D space, so marker-based motion capture systems [13] were designed to collect highly accurate locations of human joints. However, they are often expensive and impractical for recording realistic action videos. Fortunately, cost-effective depth cameras (e.g. Kinect sensor [14]) were designed to provide reliable joint locations via real-time pose estimation algorithms (e.g., [15]). Our method is general enough to be applied to various kinds of data. To extensively evaluate the proposed methods, we conducted experiments on four datasets containing examples of all three types of data: JHMDB [31], SBU [32], Berkeley MHAD [33], and NTURGB+D [21]. The information we used herein for action recognition is the locations of landmarks in all the frames. However, it is worth noting that our method is flexible and additional information such as visibility state or confidence score can be included.

The JHMDB dataset [31] is a 2D human action dataset. There are 928 clips, each clip containing between 15 and 40 frames. A clip captures only one person doing one of 21 actions. The 2D joint positions are manually annotated. There are 3 splits separating the whole dataset into training and testing set. The final result is the average of them. The sub-JHMDB is a subset of JHMDB with the full body inside the frame. The challenges are the spontaneity of the actions captured by the clips from YouTube and the loss of information due to 2D projection.

The SBU Interaction [32] is a 3D Kinect-based dataset. It has 282 clips categorized into 8 classes of two-actor interactions, and has 30 joints per frame. Its depth information suffers from self-occlusion, causing measurement errors in the estimated joint locations.

The third dataset is Berkeley MHAD dataset [33] captured by a marker-based motion capture system. It consists of 659 clips, of which 384 clips, performed by 7 actors, are used for training and 275 clips by 5 different actors are used for testing. The 3D locations of 43 joints captured by LED markers are very accurate.

The Kinect-based NTURGB+D [21] is one of the largest 3D action recognition datasets and contains 56 thousand clips of 60 classes. The large viewpoint variations and unconstrained number of actors pose considerable challenges for analysis of this dataset.

Note that the quantitative analysis was conducted on JHMDB, and all the parameters were determined by 5-fold cross validation on the training set of the first split.

## 6.2 Network configurations

Since PSFs are rich non-linear features, we adopted a single-hidden-layer linear neural network as our classifier (1-layer net also works well in preliminary experiments). The network is fully-connected and the activation of the hidden neurons is the identity function. The input dimension is decided by the PSF (i.e., **S-P-PSF**, **S-T-PSF**, **T-J-PSF**, **T-S-PSF**, or some combinations of them) and the output is a probability distribution given by a softmax layer over all the class labels in a dataset. The single hidden layer has 64 neurons. The networks are trained by stochastic gradient descent on the cross-entropy with momentum 0.7 and mini-batch size 30. The learning rate updates in accordance to $\alpha(t)=\alpha(0)\cdot\exp(-\lambda t)$ where $\alpha(0)=0.005$ $\lambda=0.005$. The maximum epoch is 300 for all experiments.

Dropconnect [72], a generalization of Dropout [73], randomly omits a proportion of connections at each training iteration. It is applied to the connections between the input and the single hidden layer for regularization. A high ratio of Dropconnect is essential to mitigate overfitting because the features herein are of very high dimension. Additionally, since the actions of some joints are highly correlated with each other, a small proportion of joints or features may already be sufficient to distinguish some actions. Based on our preliminary experiments, the Dropconnect rate is set to 0.95.

## 6.3 Data preprocessing and benchmark

We used two kinds of data augmentation. One is horizontal flipping, and the other one is adding Gaussian noise (inspired by [31]) over joint coordinates to simulate the noisy positions caused by estimation or annotation.

To cope with translation variation, we normalized the joints from world coordinate system to person-centric coordinate system by placing the center point of the body at the origin. To compensate for the biometric differences, we normalized the coordinate values to the range of [-1,1] over the entire clip. For feature normalization, each feature was divided by the maximum absolute value of the corresponding dimension and normalized to [-1, 1].

The spatial components (**S-J**, **S-P-PSF**, and **S-T-PSF**) are calculated for each frame. To obtain a fixed-length input to the network, we uniformly sampled $M$ (in this paper, $M=10$) frames from each clip. As the signature has a fixed dimension under length variation, the temporal features (**T-J-PSF** and **T-S-PSF**) are extracted from all the frames without subsampling. Our baseline method is using **S-J**, the $d$-dimensional coordinate values of all $N$ joints. This leads to $MNd$-dimensional feature set, for which we obtained a validation error rate of $57.54\pm3.26\%$.

## 6.4 Investigation of the spatial features

As described in section 4.1.1 and 5.1, by pose disintegration with $m=2$ and $m=3$, we constructed all the joint pairs and triples as localized paths for **S-P-PSF** and **S-T-PSF** respectively. The error rates on the validation set obtained by these feature sets are shown in Table 2 and Table 3. The performance improves when higher terms of the signature are considered, but the improvements tend to be negligible and the variance increases when the dimension of the feature grows exponentially with the signature level $n$. For the joint pairs, a suitable truncation level $n_{SP}$ is 2 or 3, while for the joint triples, the level $n_{ST}$ needs to be as high as 3 or 4, which suggests the choice of $n$ should depend on $m$. We refer the reader to [94] which discusses the relationship among $m$, $n$, and the path dimension $d$ from the view of path recovery. For the following experiments, we chose to fix $n_{SP}=2$, $n_{ST}=3$.

TABLE 2
EFFECT OF DIFFERENT SIGNATURE LEVELS ON THE PERFORMANCE OF **S-P-PSF**

| Type of subpaths | Signature level $n_{SP}$ | Feature dim. | Error rate (%) |
|---|---|---|---|
| Joint Pairs | 1 | 2100 | 32.79±4.49 |
| | 2 | 6300 | 25.41±4.55 |
| | 3 | 14700 | **24.10±5.65** |
| | 4 | 31500 | 24.10±5.72 |

TABLE 3
EFFECT OF DIFFERENT SIGNATURE LEVELS ON THE PERFORMANCE OF **S-T-PSF**

| Type of subpaths | Signature level $n_{ST}$ | Feature dim. | Error rate (%) |
|---|---|---|---|
| Joint Triples | 1 | 9100 | 43.93±2.87 |
| | 2 | 27300 | 32.46±3.26 |
| | 3 | 63700 | 26.39±3.99 |
| | 4 | 136500 | **24.75±4.79** |
| | 5 | 282100 | 23.77±6.41 |
| | 6 | 573300 | 25.24±6.44 |

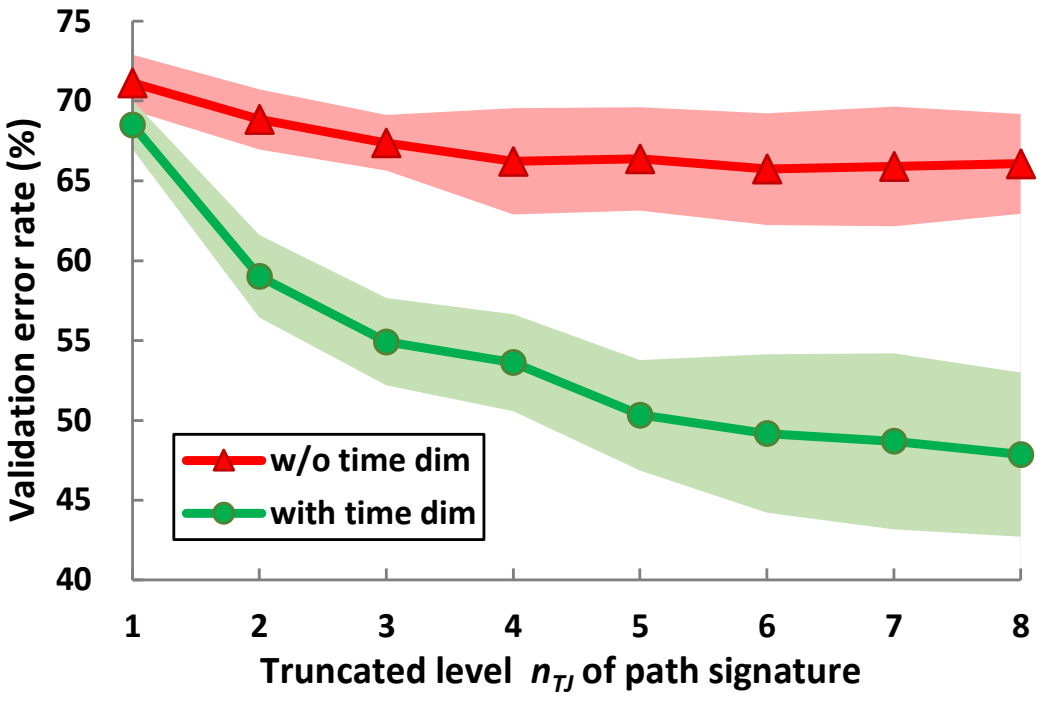

Fig. 6. Comparison of **T-J-PSF** w/ and w/o using time-incorporated paths. The colored areas are the error bands.

## 6.5 Investigation of the temporal features

### *6.5.1 Investigation of T-J-PSF*

First, we investigated the effect of the time-incorporated transformation and the truncation level $n_{TJ}$ of the **T-J-PSF**. As shown in Fig. 6, if the truncation level $n_{TJ}$ (the horizontal axis) is 1, adding a time dimension (the green plot) only improves the performance a little. This is because the first level term related to the time dimension is only the duration of the action. When $n_{TJ}$ increases, the performance improvements of using time-incorporated PSF are more significant, showing the effectiveness of the time-incorporated path transformation. As to the truncation level, when $n_{TJ}$ increases, the results have lower bias together with gradually higher variance, so a tradeoff is required. Here, $n_{TJ}=5$ is a good choice.

In addition, we investigated the effect of the signature of the time-incorporated path at different frame rates. We artificially increased the frame rate by interpolating additional frames at random time stamps of the original clips. Bodies of the additional frames were copied from those of their adjacent frames. On the other hand, we decreased the frame rate by random subsampling. The networks were trained using the training clips at original frame rate (30fps) and tested 10 times using artificial validation clips at each of the frame rates ranging from 6fps to 90fps in 6fps steps. As shown in Fig. 7, when the frame rate increases from 30fps to 90fps, the error rates of using the time-independent part (TI) stay the same, while those of using the time-dependent part (TD) raise rapidly. It demonstrates the TI (i.e. the signature of original path) is invariant under time reparameterization while the TD is very sensitive to speed variation. The larger the signature level $n$, the more sensitive the TD is to speed variation. Similarly, in the other direction, when the frame rate decreases from 30fps to 6fps, the influence to TD is far more significant than that to TI, showing the tolerance of TI under missing frames.

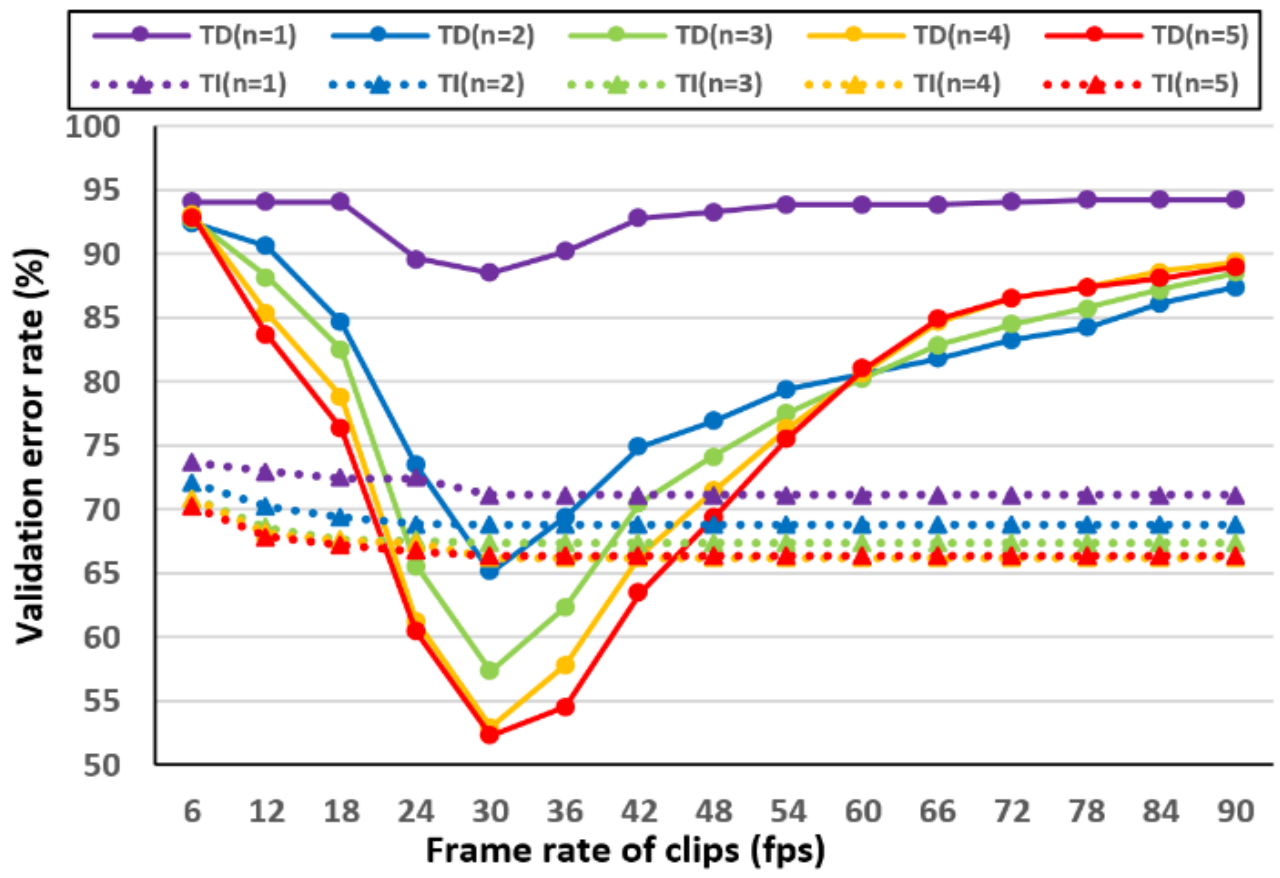

Fig. 7. Sensitivity of the time-dependent and time-independent part of the time-incorporated PSF to different frame rates.

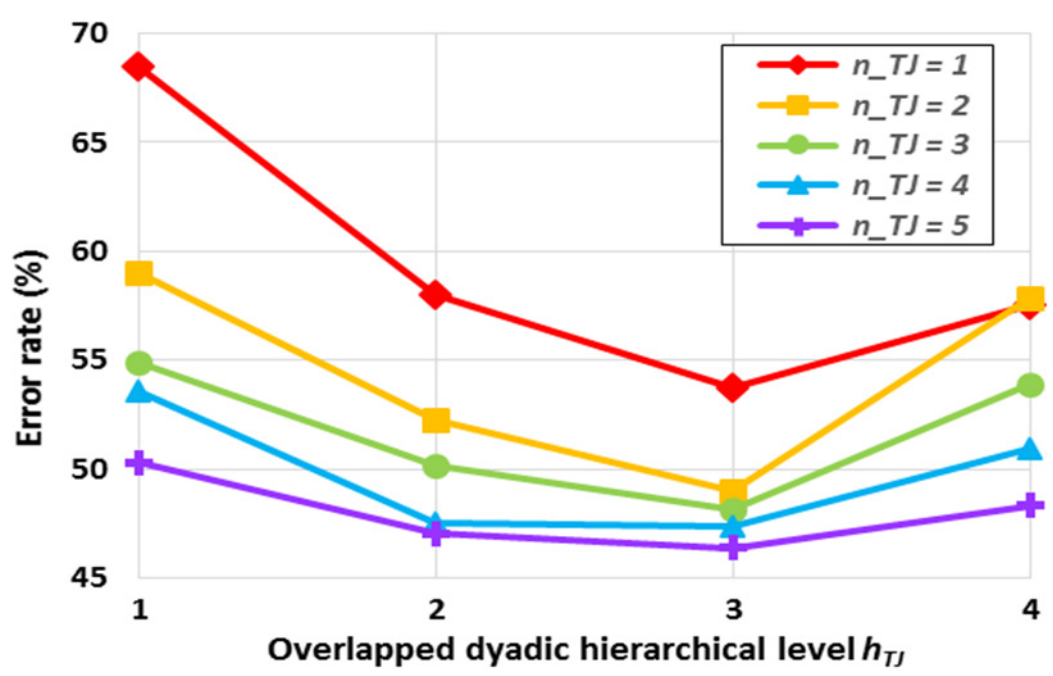

Fig. 8. Comparison of **T-J-PSF** with different dyadic hierarchical level $h_{TJ}$ and different truncation level $n_{TJ}$ of signature.

If we replace the PSF with the overlapping DPSF, then an appropriate hierarchical level $h_{TJ}$ needs to be chosen. As shown in Fig. 8, in terms of performance, the low-level (e.g., $n_{TJ} = 2$) overlapping DPSFs over the hierarchical intervals (e.g., $h_{TJ} = 3$) often outperform the high-level (e.g., $n_{TJ} = 5$) PSFs over the whole interval ($h_{TJ} = 1$), which shows the efficiency of using temporal disintegration. However, when the disintegrated paths are too fragmented to avoid being dominated by local noise (e.g., when $h_{TJ} > 3$), the additional features are harmful. We thus fixed $h_{TJ} = 3$. Another observation is that the improvements from $h_{TJ} = 1$ to $h_{TJ} = 3$ become less significant along with the increasing $n_{TJ}$, demonstrating a trend that the high-level PSF and low-level DPSF yield similar information eventually.

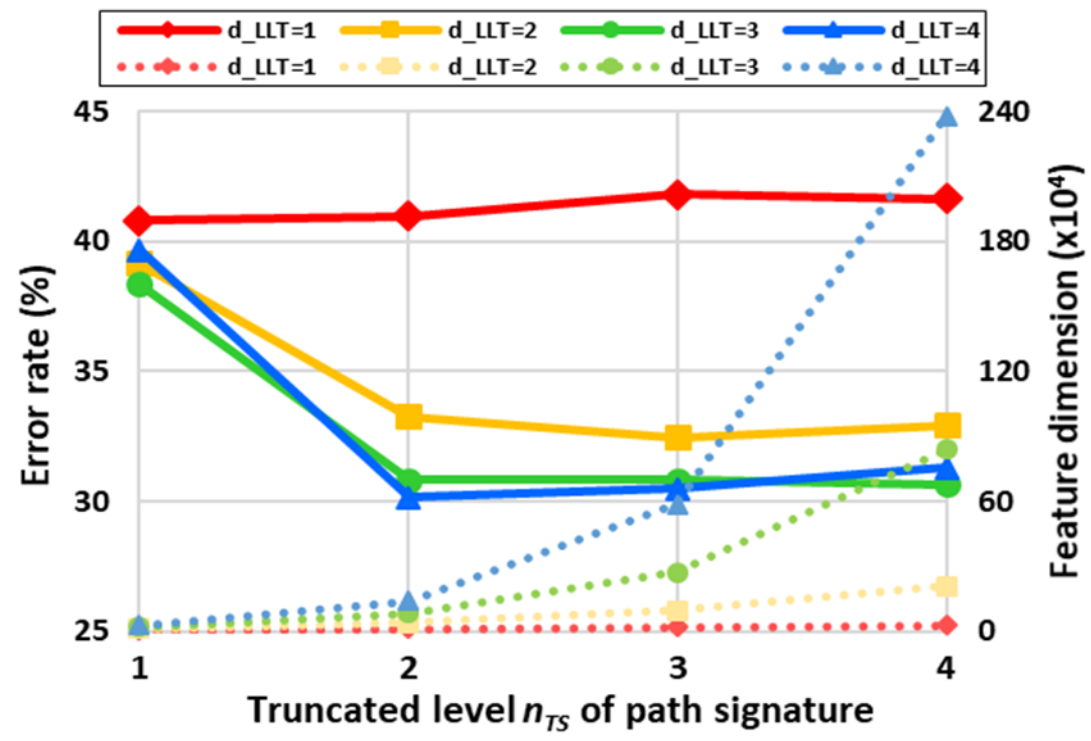

Fig. 9. Comparison of the error rates (solid) and the feature dimensions (dashed) of using **T-S-PSF** with different dimensions $d_{LLT}$ of lead-lag paths and different truncation level $n_{TS}$ of signature.

TABLE 4
THE ABLATION STUDY OF OUR FEATURES ON JHMDB

| Ex. # | S-J | S-P-PSF | S-T-PSF | T-J-PSF | T-S-PSF (S-P)* | T-S-PSF (S-T)* | Accuracy (%) |
|---|---|---|---|---|---|---|---|
| 1 | O | | | | | | 48.9 |
| 2 | O | O | | | | | 68.4 |
| 3 | O | | O | | | | 68.7 |
| 4 | O | O | O | | | | 69.2 |
| 5 | O | | | O | | | 62.0 |
| 6 | O | O | O | O | | | 73.5 |
| 7 | O | O | | O | O | | 79.1 |
| 8 | O | | O | O | | O | 78.3 |
| 9 | O | O | O | O | O | O | 80.4 |

** S-P (S-T) means the corresponding temporal features are only on the base of spatial joint pairs (spatial joint triples).*

### 6.5.2 Investigation of T-S-PSF

Regarding the PSF derived from the evolution of the spatial context (**T-S-PSF**), two factors were evaluated: the dimension $d_{LLT}$ of the lead-lag path and the truncation level $n_{TS}$ of the signature. As shown in Fig. 9, the results improve when a higher dimension $d_{LLT}$ of the lead-lag path is adopted, but the marginal improvement is less obvious when $d_{LLT} \geqslant 3$. For the truncation level $n_{TS}$, the improvements are significant from $n_{TS} = 1$ to $n_{TS} = 2$, but they are negligible when $n_{TS} > 2$. The dashed lines in Fig.9 show the trends of feature dimension under different parameters. By making a tradeoff between model complexity and performance, we fixed $d_{LLT} = 3$, $n_{TS} = 2$.

By using the overlapping DPSF instead of PSF, the validation error rates are 30.82 ± 7.00%, 26.07 ± 6.12%, 26.39 ± 5.51%, and 26.07 ± 5.23%, when the hierarchical level $h_{TS}$ is 1, 2, 3, and 4 respectively. Thus, we fixed $h_{TS}$ to 3.

## 6.6 Ablation study

For the ablation study of our features on the JHMDB [31], we used the parameter setting for each feature based on the foregoing analysis. We retrained the network using the whole training set (including the validation set) and took the final result as the average of the three splits. The results are shown in Table 4. We can see that adding the spatial PSF (Ex. 4) to the baseline (Ex. 1) gives an improvement of about 20%, and further adding the temporal PSF (Ex. 9) contributes an additional 10%. The spatial context may be alternative between joint pairs and joint triples, for example Ex. 2 vs. Ex. 3, or Ex. 7 vs. Ex. 8, but they are complementary as shown in Ex. 4 and Ex. 9.

Applying the invisibility-reset transformation to all the paths before taking the temporal signatures allow us to remove all the spatial components **S-J**, **S-P-PSF**, and **S-T-PSF**, while obtain the same accuracy as that of Ex. 9.

Also, we evaluated the method which directly takes all the evolving $N$ landmarks in $\mathbb{R}^d$ as a $Nd$-dimensional path for signature extraction. Together with **S-J**, it achieves 55.0% in accuracy. The dimension of this PSF is $\varphi'(Nd, n) = 838{,}230$ when $n = 4$ and will be impractical when $n > 4$.

### 6.7 Comparison with the state-of-the-art methods

To achieve state-of-the-art results, we adopted the best settings of parameters from the foregoing analysis. For the JHMDB dataset [31], the results were given in the previous subsections. For the other three datasets, we followed the evaluation criteria in [22].

#### 6.7.1 Comparison over small datasets

For the JHMDB dataset, previous state-of-the-art methods are high-level pose feature (HLPF) [31] and its modified version (i.e. Novel HLPF [44]), dense trajectory features [74] encoded by Fisher vectors [75], and the pose-based CNN features (P-CNN) [56]. As shown in Table 5, our method, which uses only the joint locations, achieve better performance than the P-CNN which requires additional RGB information. Further, our method manages the high degree of nonlinearity, and outperforms other methods using hand-designed features like HLPF.

Moreover, we used the off-the-shelf pose estimation called Alphapose (with Poseflow) [88] to get a set of 17 estimated joints from the RGB videos of the sub-JHMDB dataset, and then trained and tested the network using the estimated poses. By using only location information, our test accuracy is 68.2%, which outperforms that of P-CNN [56] (66.8%), PA-AP [89] (61.5%), JointAP [90] (61.2%), or HLPF [31] (54.1%). As an example of the flexibility of our method on additional clues, taking the confidence scores from the pose estimation as an additional dimension of landmarks raises the accuracy rate to 75.7%. However, a gap of accuracy still exists between using estimated poses and ground truth poses (84.23% by ours).

For the SBU Interaction dataset, the two human bodies are regarded as one united articulated system with a total of 30 joints in 3D. As shown in Table 6, the proposed method using PSF significantly outperforms the other skeleton-based methods including many RNN-based or LSTM-based ones. Aside from the accuracy, the interpretable PSF could facilitate further understanding of interactions. For the Berkeley MHAD dataset, we achieve the same accuracy (100%) as the state-of-the-art methods shown in Table 7, showing the effectiveness of PSF for recognizing actions with accurate joint locations.

TABLE 5
COMPARISON OF METHODS ON JHMDB USING GROUNDTRUTH

| Method | Accuracy (%) |
| --- | --- |
| DT-FV [74] | 65.9 |
| P-CNN [56] | 74.6 |
| HLPF [31] | 76.0 |
| Novel HLPF [44] | 79.6 |
| Path Signature (Ours) | **80.4** |

TABLE 6
COMPARISON OF METHODS ON SBU DATASET

| Method | Accuracy (%) |
| --- | --- |
| Yun et al., [32] | 80.3 |
| Ji et al., [76] | 86.9 |
| CHARM [77] | 83.9 |
| HBRNN [19] (reported by [18]) | 80.4 |
| Deep LSTM (reported by [18]) | 86.0 |
| Co-occurrence LSTM [18] | 90.4 |
| STA-LSTM [78] | 91.5 |
| ST-LSTM-Trust Gate [22][23] | 93.3 |
| SkeletonNet [79] | 93.5 |
| GC-Attention-LSTM [86] | 94.1 |
| Path Signature (Ours) | **96.8** |

TABLE 7
COMPARISON OF METHODS ON MHAD

| Method | Accuracy (%) |
| --- | --- |
| Vantigodi et al. [80] | 96.1 |
| Ofli et al. [49] | 95.4 |
| Vantigodi et al. [81] | 97.6 |
| Kapsouras et al. [82] | 98.2 |
| HBRNN [19] | 100 |
| ST-LSTM-Trust Gate [22][23] | 100 |
| Path Signature (Ours) | 100 |

#### 6.7.2 Comparison over large-scale datasets

We also conducted experiments on the cross-subject and cross-view tasks of the large-scale NTURGB+D data.

For normalization, we applied the same 3D rotation and scaling as those in [21], so the body in the first frame faces the camera directly and those in the following frames are compensated accordingly. Since in this dataset different actions contain different number of detected actors, we applied a two-stage classification. The first stage is a binary classifier separating the actions into two types: 1-body or 2-body actions, then the second stage is the corresponding classifier (1-body or 2-body classifier) for each type. The supervised label of the binary classification at the first stage can be found by going through all the training samples and calculating the average number of actors in each action class. The range of the numbers is [1.02, 1.06] for the first 49 classes which are annotated as 1-body actions, while the range is [1.87, 2.04] for the remaining 11 classes which are annotated as 2-body actions.

TABLE 8
ACCURACY (%) OF THE TWO-STAGE CLASSIFICATION ON NTURGBD

| Task | The 1st stage | The 2nd stage | | Final |
|---|---|---|---|---|
| | | 1-body | 2-body | |
| Cross-subject | 99.2 | 75.7 | 91.9 | 78.3 |
| Cross-view | 99.3 | 82.5 | 94.4 | 86.1 |

TABLE 9
COMPARSION OF METHODS ON NTURGB+D

| Method | Cross-subject | Cross-view |
|---|---|---|
| Dynamic Skeletons [84] | 60.2 | 65.2 |
| HBRNN [19] | 59.1 | 64.0 |
| Part-aware LSTM [21] | 62.9 | 70.3 |
| ST-LSTM-Trust Gate [22][23] | 69.2 | 77.7 |
| STA-LSTM [78] | 73.4 | 81.2 |
| SkeletonNet [79] | 75.9 | 81.2 |
| Joint Distance Maps [85] | 76.2 | 82.3 |
| GC-Attention-LSTM [86] | 74.4 | 82.8 |
| Path Signature (Ours) | **78.3** | **86.1** |

Before feature extraction, we ranked all the detected actors in each clip based on the magnitudes of their movements. Then, for the 1-body classifier, features were extracted from the most active actor. For the first-stage binary classifier and the 2-body classifier, the two most active actors were regarded as one evolving object; this means we ended up having twice the number of joints per frame (i.e., 50 joints per frame). If a body was missing in the entire clip, the coordinates of this body were set to 0; if a body was missing in some medial frames, its coordinates were filled in using cubic spline interpolation [83].

The final results were given by two-stage classification as shown in Table 8. Table 9 shows that our method outperforms other methods including RNN-based ones, showing the effectiveness of the proposed features. The current state-of-the-art method [92] exploits skeletal structure as a graph for CNN and [96] fuses the scores from RNNs and CNNs.

### 6.8 Toward understanding of human actions

The interpretable geometric properties of PSF facilitate the understanding of human actions. By using a linear classifier, the importance of each feature to each action class can be evaluated by the product of the two-layer weight matrices. For each class of sub-JHMDB, we ranked the joint pairs/triples according to the average over the weights connecting the features of joint groups and the corresponding class label. The top-3 joint pairs/triples for spatial and temporal features are shown in Fig. 10. The spatial ones often emphasize static constraints while the temporal ones highlight dynamic variations. Notice that many top pairs/triples are physically non-local, which demonstrates the effectiveness of the pose disintegration method.

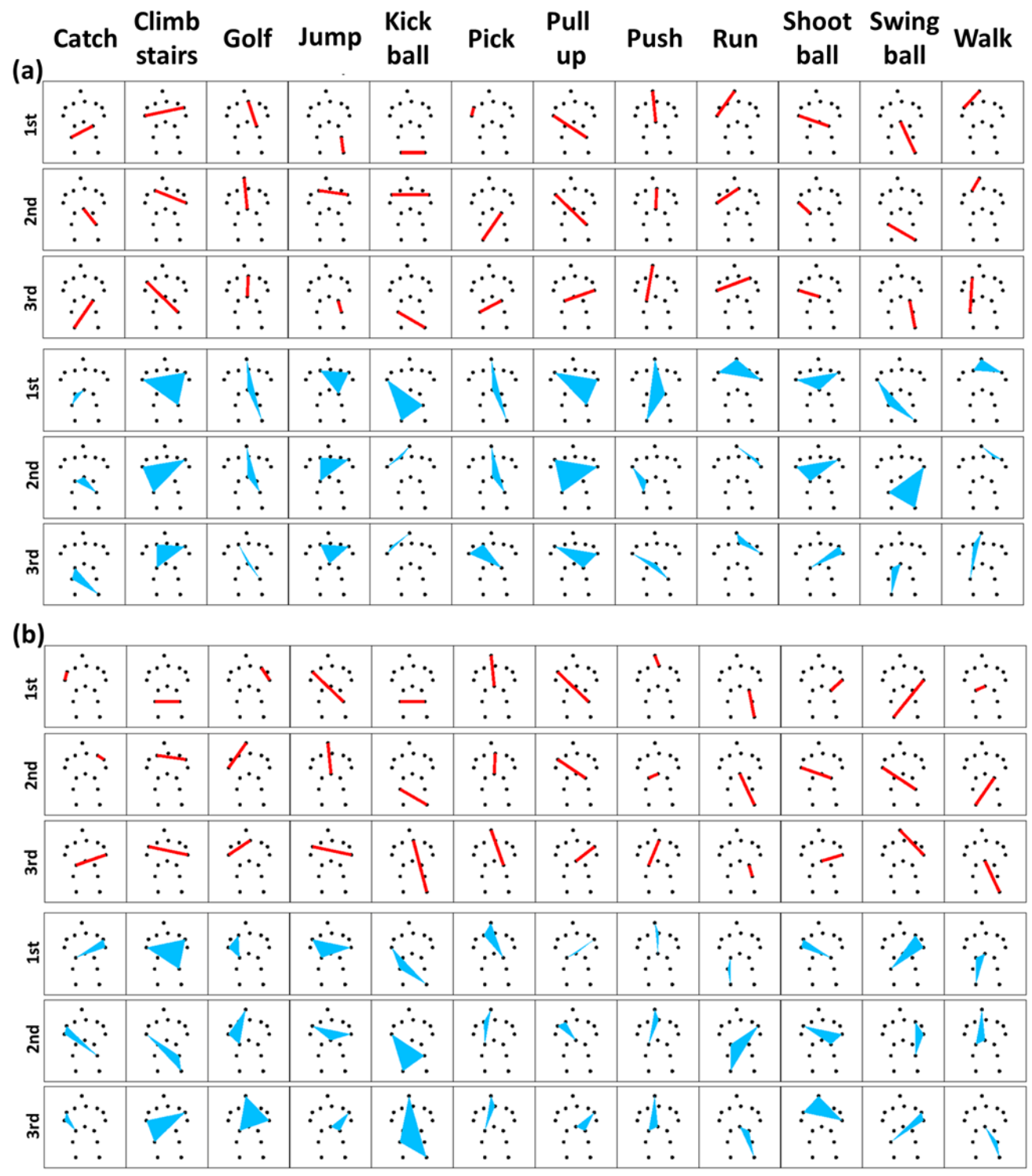

Fig. 10. Top-3 most important joint pairs/triples for (a) spatial features and (b) temporal features based on our linear network.

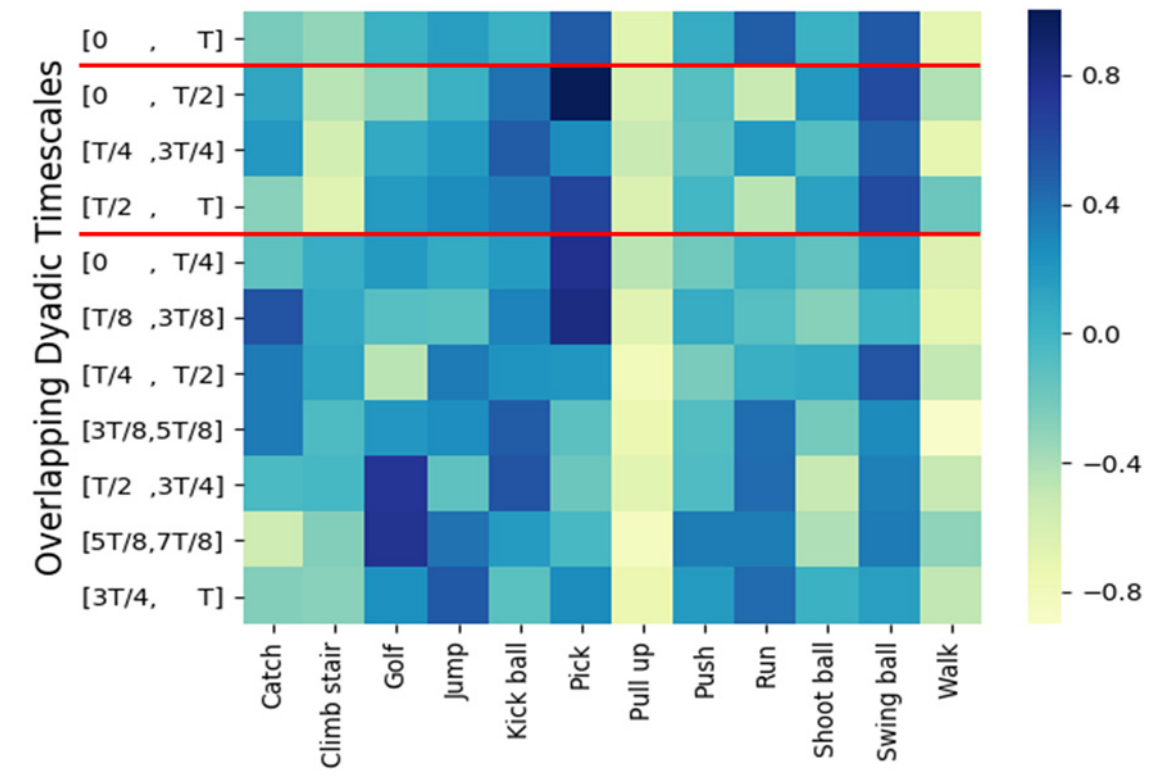

Fig. 11. The important timescales and time periods for the actions on sub-JHMDB. The darker in color, the more important it is.

Moreover, by using temporal disintegration ($h = 3$), we can evaluate the importance of different timescales and time intervals. As shown in Fig. 11, discriminative motions often appear in various intervals of finer timescales, e.g., the start of “catch” or “pick”, the middle of “kick ball” or “swing ball”, and the end of “golf” or “jump”.

## 7 CONCLUSIONS

In this paper, we refined the path signature as a robust, nonlinear, and interpretable feature for landmark-based data. Path disintegrations and transformations are proposed to improve the effectiveness and efficiency of signature features. Based on these, we designed and built the signature-based spatio-temporal representation of action sequences. Experimental results show that using our feature set, a linear shallow fully-connected neural network achieves comparable results to advanced methods including CNN-based and RNN-based ones.

For future work, one could reduce the size of the representation of the body or feature set based on our analysis and understanding of human actions. It would also be interesting to integrate our landmark-based representation with other informative cues (e.g., appearance) to improve the performance of HAR. Moreover, our method is general enough for other landmark-based objects where the given information in each landmark can be diverse.

## ACKNOWLEDGMENT

T. L. and H. N. are supported by the Alan Turing Institute under the EPSRC grant EP/N510129/1. This work was supported by ERC advanced grant ESig (no. 291244). C. S. was supported in part by ERC advanced grant ALLEGRO. This work was supported in part by the Alexander von Humbolt Foundation. W. Y. and L. J. are supported in part by NSFC (Grant no.: 61472144, 61673182), the National Key Research & Development Plan of China (no. 2016YFB1001405), GD-NSF (no. 2017A030312006), GDSTP (Grant no.: 2015B010101004, 2015B010130003), GZSTP (no. 201607010227). W. Y. is supported by a Royal Society Newton International Fellowship.

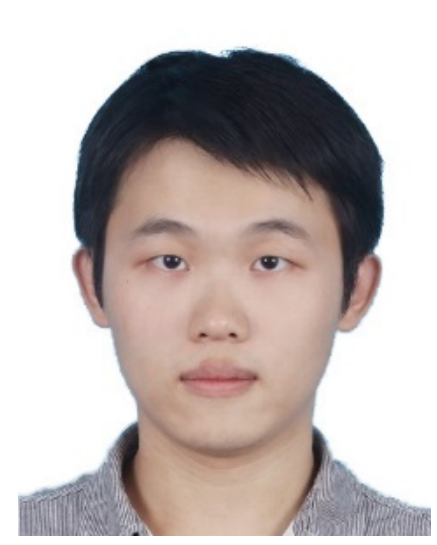

**Weixin Yang** is a post-doctoral research assistant visiting at the Mathematical Institute of Oxford University. Prior to this He was a PhD student in South China University of Technology from 2013/09 to 2018/09. He was awarded the Royal Society Newton International Fellowship in 2018. His research interests include Computer vision, Machine Learning and sequential data analysis.

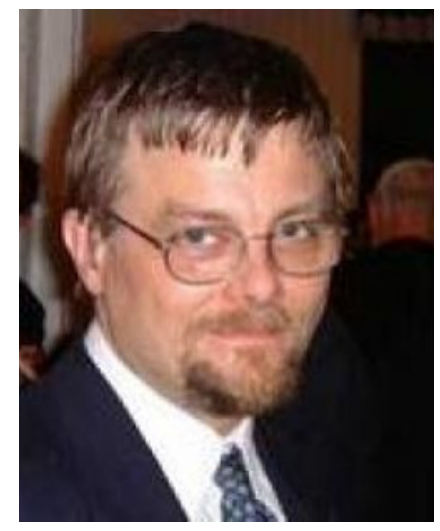

**Terry Lyons** is the Wallis Professor of Mathematics of Oxford University. He was a founding member (2007) of, and then Director (2011-2015) of, the Oxford Man Institute of Quantitative Finance. He was the Director of the Wales Institute of Mathematical and Computational Sciences (WIMCS; 2008-2011). Lyons came to Oxford in 2000 having previously been Professor of Mathematics at Imperial College London (1993-2000), and before that he held the Colin Maclaurin Chair at Edinburgh (1985-93). His research interests are focused on Rough Paths, Stochastic Analysis, and Applications. He is also interested in developing mathematical tools that can be used to effectively model and describe high dimensional systems that exhibit randomness. He was President of the UK Learned Society for Mathematics, the London Mathematical Society (2013-2015)

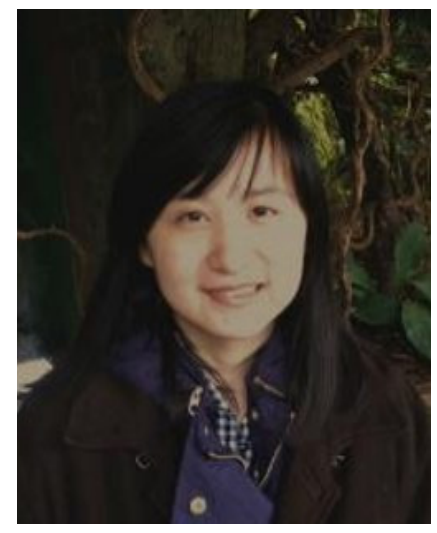

**Hao Ni** is a senior lecturer in financial mathematics at UCL since September 2016. Prior to this she was a visiting postdoctoral researcher at ICERM and Department of Applied Mathematics at Brown University from 2012/09 to 2013/05 and continued her postdoctoral research at the Oxford-Man Institute of Quantitative Finance until 2016. She finished her PhD in mathematics in 2012 under the supervision of Professor Terry Lyons at University of Oxford.

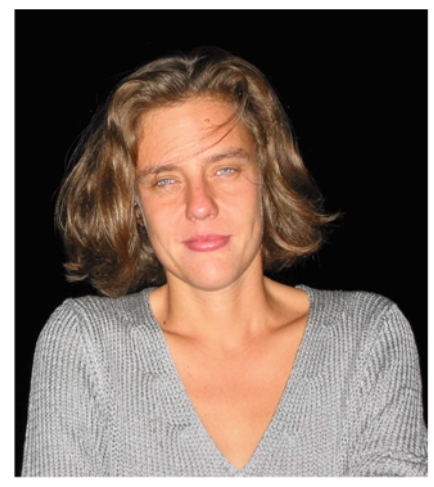

**Cordelia Schmid** holds a M.S. degree in Computer Science from the University of Karlsruhe and a Doctorate, also in Computer Science, from the Institut National Polytechnique de Grenoble (INPG). Her doctoral thesis received the best thesis award from INPG in 1996. Dr. Schmid was a post-doctoral research assistant in the Robotics Research Group of Oxford University in 1996–1997. Since 1997 she has held a permanent research position at INRIA Grenoble Rhone-Alpes, where she is a research director and directs an INRIA team. Dr. Schmid is the author of over a hundred technical publications. She has been an Associate Editor for IEEE PAMI (2001–2005) and for IJCV (2004–2012), editor-in-chief for IJCV (2013—), a program chair of IEEE CVPR 2005 and ECCV 2012 as well as a general chair of IEEE CVPR 2015 and ECCV 2020. In 2006, 2014 and 2016, she was awarded the Longuet-Higgins prize for fundamental contributions in computer vision that have withstood the test of time. She is a fellow of IEEE. She was awarded an ERC advanced grant in 2013, the Humbolt research award in 2015 and the Inria & French Academy of science Grand Prix in 2016. She was elected to the German National Academy of Sciences, Leopoldina, in 2017.

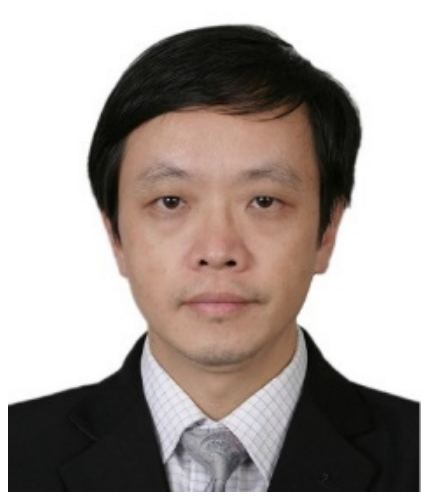

**Lianwen Jin** received the B.S. degree from the University of Science and Technology of China, Anhui, China, and the Ph.D. degree from the South China University of Technology, Guangzhou, China, in 1991 and 1996, respectively. He is currently a Professor with the School of Electronic and Information Engineering, South China University of Technology. He is the author of more than 100 scientific papers. Dr. Jin was a recipient of the award of New Century Excellent Talent Program of MOE in 2006 and the Guangdong Pearl River Distinguished Professor Award in 2011. His research interests include image processing, handwriting analysis and recognition, machine learning, deep learning, and intelligent systems.